\documentclass{article}
\usepackage{arxiv}

\usepackage{amsmath}
\usepackage{amsthm}
\usepackage{amssymb}
\usepackage{graphicx}
\title{Moving Past Single Metrics: Exploring Short-Text Clustering Across Multiple Resolutions}

\author{
 Justin K. Miller \\
  School of Physics\\
  University of Sydney\\
  Camperdown, NSW 2006 \\
  \texttt{justin.k.miller@sydney.edu.au} \\
   \And
 Tristram J. Alexander \\
  School of Physics\\
  University of Sydney\\
  Camperdown, NSW 2006 \\
  \texttt{tristram.alexander@sydney.edu.au} \\
}

\begin{document}
\maketitle


\begin{abstract} 
Cluster number is typically a parameter selected at the outset in clustering problems, and while impactful, the choice can often be difficult to justify. Inspired by bioinformatics, this study examines how the nature of clusters varies with cluster number, presenting a method for determining cluster robustness, and providing a systematic method for deciding on the cluster number.  The study focuses specifically on short-text clustering, involving 30,000 political Twitter bios, where the sparse co-occurrence of words between texts makes finding meaningful clusters challenging.  A metric of proportional stability is introduced to uncover the stability of specific clusters between cluster resolutions, and the results are visualised using Sankey diagrams to provide an interrogative tool for understanding the nature of the dataset.  The visualisation provides an intuitive way to track cluster subdivision and reorganisation as cluster number increases, offering insights that static, single-resolution metrics cannot capture.  The results show that instead of seeking a single ‘optimal’ solution, choosing a cluster number involves balancing informativeness and complexity.

\end{abstract}

\section{Introduction}

When faced with a large collection of documents, it can be challenging to determine the nature of the topics the documents are covering.  One way to approach this problem is to group documents together by theme, thus making it easier to understand the nature of the overall collection.  This grouping of documents can be framed as a clustering problem, in which the goal is to deduce the categories within the data~\cite{van2023white}.  Typically this categorisation would proceed by seeking to maximise the insight provided by the categorisations, while minimising the cognitive load required to determine

which document goes into which category~\cite{roschPrinciplesCategorization1978}.  The effectiveness of clustering is thus contingent on the balance between the simplicity of the categories and the usefulness of the information the categories convey~\cite{hennig2019cluster}.

The challenge in text clustering therefore lies in finding a balance between creating clusters that are informative enough to capture meaningful distinctions in the data while remaining interpretable and easy to apply. This study examines how varying the number of clusters affects this trade-off between informativeness and interpretability, highlighting that there is no universally optimal clustering solution, but rather context-dependent cluster structures that best balance these competing factors.

Cluster validation indices provide a multidimensional assessment that is essential for selecting the number of clusters in line with specific analysis goals \cite{hennig2019cluster}. However, the choice of validation metric can significantly impact the determination of an `ideal' cluster number, as different metrics emphasise different clustering characteristics \cite{akhanliComparingClusteringsNumbers2020}.  For example, metrics that prioritise within-cluster homogeneity, such as measures of low within-cluster distance, may lead to a preference for a higher number of clusters, as they aim to minimize internal variance. Conversely, metrics that favor between-cluster separation, such as separation indices, can favor a smaller number of clusters by emphasizing distinct boundaries between groups. Selecting an appropriate clustering solution therefore requires balancing these metrics based on the dataset's characteristics, the user's objectives, and broader considerations such as interpretability \cite{grimmer2021machine}. 
While achieving high performance metrics that align with pre-established labels or metadata is often emphasised, clustering's ultimate objective should be to organise data in ways that are practically useful, rather than merely identifying ``correct'' categories \cite{pedryczFuzzyClusteringKnowledgebased2004, lakhawatNovelClusteringAlgorithm2016, krausMultiobjectiveSelectionCollecting2011, krishnapuramPossibilisticApproachClustering1993, akhanliComparingClusteringsNumbers2020}.

When evaluating clustering solutions, it is possible to adopt either a single-level or a multi-level perspective. Single-level evaluation focuses on assessing the quality of clusters at a fixed number of clusters ($K$), treating the solution as static and independent of other potential resolutions. Methods such as coherence~\cite{stevensExploringTopicCoherence2012} and silhouette scores~\cite{rousseeuwSilhouettesGraphicalAid1987} represent this single-level approach: they yield valuable insights into cluster compactness and separability but only for one chosen $K$. In contrast, a multi-level approach considers how cluster structures evolve as $K$ changes, making it possible to determine whether clusters persist across multiple resolutions or simply arise at a particular $K$. By capturing these dynamic relationships and structural transitions, a multi-level perspective can reveal whether clusters represent enduring, meaningful patterns in the data or whether they emerge and vanish as the clustering granularity shifts.

A multi-level approach therefore not only provides greater insight into the underlying data structure but also leads to clustering solutions that more closely reflect the dataset’s characteristics and the user’s objectives. Developing methodologies that incorporate multi-level analysis thus offers a more comprehensive and nuanced validation process.

In this work, the evolution and stability of clusters in short-text data are examined across a range of resolutions. The methodology incorporates multi-level stability assessments using Adjusted Mutual Information (AMI) and introduces a novel metric, Proportional Stability, alongside Sankey diagram visualisations to illustrate cluster transitions. These tools capture both the robustness of clustering outcomes and the deeper structural changes that occur as the number of clusters increases. By highlighting the importance of examining cluster dynamics across resolutions, this work advances clustering validation methodologies and supports more interpretable, contextually meaningful results in short-text clustering applications.

\section{Related Work}
\subsection*{Stability of clusters}
Most clustering methods involve finding clusters after a random initialisation process~\cite{ahmedKmeansAlgorithmComprehensive2020, hosseiniAlternativeEMGaussian2020}.  
This stochastic element in the initialisation means that the resulting clusters may be different with different initialisations. Cluster stability measures how consistently an algorithm finds the same clusters within a dataset across the different random starting points.  Cluster stability is therefore a measure which may be used to assess the robustness and reliability of clustering methods, with the assumption being that if there is some underlying structure in the data, the algorithm should be able to recover this consistently.

Cluster stability is central in the field of bioinformatics, where the high-dimensional nature of the data, coupled with biological variation, makes stability an indicator that a clustering algorithm has captured biologically meaningful clusters (e.g., distinct cell types) \cite{kiselevChallengesUnsupervisedClustering2019}. 
Importantly, in bioinformatics, the notion of stability is tied to the biological reproducibility of the findings: when clusters correspond to real biological phenomena, they should persist across different algorithms, initialisations, and even perturbations to the dataset \cite{handlComputationalClusterValidation2005, ronanAvoidingCommonPitfalls2016}. To test how stable the clusters found by an algorithm are, one must consider both the variance in the data, and also the variance in the dimensions used i.e. ensuring that the clusters are not just forming as a result of variance within a few data points, or from a few variables \cite{yu2022benchmarking}.  This work will draw on the approaches of bioinformatics, but applied to text analysis.

A key difference between bioinformatics and text clustering lies in the interpretability of the clusters themselves. While biological clusters often have clear interpretations based on known cell types or functions \cite{qiClusteringClassificationMethods2020}, the interpretability of text clusters relies heavily on human judgment \cite{JMLR:v18:17-069}. 
This adds another dimension to the utility of stability. In text clustering, stability might suggest that the algorithm has found coherent groupings, but it doesn’t guarantee that these groupings are semantically meaningful or useful for the task at hand \cite{ahmedKmeansAlgorithmComprehensive2020}. Hence, an additional human interpretability step is often needed to evaluate whether the stable clusters align with meaningful categories or concepts \cite{kuncheva2006evaluation}.

Ultimately, stability serves as a valuable criterion for identifying meaningful clusters across fields, but the interpretation of stability can vary. In bioinformatics, stable clusters are likely to correspond to real, reproducible biological phenomena \cite{yu2022benchmarking}. In text data, stable clusters may reflect coherent patterns, but assessing their utility requires further interpretability checks, often involving human evaluation \cite{lord2017using}. Stability, in this sense, acts as a precursor to, but not a guarantee of, meaningfulness \cite{kuncheva2006evaluation}.

\subsection*{Gaussian Mixture Model Clustering}

This work uses Gaussian Mixture Models (GMMs) to find clusters in a high dimensional text embedding space.  GMMs are a probabilistic approach to clustering that assume the data are generated from a mixture of several Gaussian distributions \cite{21GaussianMixture}. Each Gaussian component represents a cluster, and the algorithm assigns probabilities of membership for each data point across these clusters, making GMM a soft clustering technique. This section outlines the theoretical foundation and practical implementation of GMM clustering \cite{mit_ml_notes_2015, JMLR:v25:23-1245, ghosh2018emnotes}.

The GMM assumes that the dataset $\mathbf{X} = \{\mathbf{x}_1, \mathbf{x}_2, \dots, \mathbf{x}_n\}$ is generated from $k$ Gaussian distributions, where each data point $\mathbf{x}_i \in \mathbb{R}^d$ belongs to a particular cluster with a certain probability.  These parameters are generally estimated through the use of the Expectation-Maximization (EM) algorithm (E-step) \cite{hosseiniAlternativeEMGaussian2020}, This paper uses the scikit learn implementation of GMM, so uses this particular algorithm \cite{21GaussianMixture}. In the E-step, the algorithm calculates the posterior probabilities (responsibilities) that each data point $\mathbf{x}_i$ belongs to each Gaussian component $j$, denoted by $\gamma(z_{ij})$, where $z_{ij}$ is the latent variable indicating the cluster membership of $\mathbf{x}_i$:

\begin{equation}
\gamma(z_{ij}) = \frac{\pi_j \, \mathcal{N}(\mathbf{x}_i \mid \boldsymbol{\mu}_j, \boldsymbol{\Sigma}_j)}{\sum_{l=1}^{k} \pi_l \, \mathcal{N}(\mathbf{x}_i \mid \boldsymbol{\mu}_l, \boldsymbol{\Sigma}_l)}.
\label{eq:e-step}
\end{equation}

\textbf{Definitions of Terms:}

\begin{itemize}
    \item \(\gamma(z_{ij})\): The posterior probability (responsibility) that data point \(\mathbf{x}_i\) belongs to the \(j\)-th Gaussian component.
    \item \(\pi_j\): The mixing coefficient (prior probability) for the \(j\)-th Gaussian component, satisfying \(\sum_{j=1}^{k} \pi_j = 1\) and \(0 \leq \pi_j \leq 1\).
    \item \(\mathcal{N}(\mathbf{x}_i \mid \boldsymbol{\mu}_j, \boldsymbol{\Sigma}_j)\): The multivariate Gaussian probability density function evaluated at \(\mathbf{x}_i\) with mean \(\boldsymbol{\mu}_j\) and covariance matrix \(\boldsymbol{\Sigma}_j\). It is defined as:
    \begin{equation}
    \mathcal{N}(\mathbf{x}_i \mid \boldsymbol{\mu}_j, \boldsymbol{\Sigma}_j) = \frac{1}{(2\pi)^{d/2} |\boldsymbol{\Sigma}_j|^{1/2}} \exp\left( -\frac{1}{2} (\mathbf{x}_i - \boldsymbol{\mu}_j)^\top \boldsymbol{\Sigma}_j^{-1} (\mathbf{x}_i - \boldsymbol{\mu}_j) \right),
    \end{equation}
    where:
    \begin{itemize}
        \item \(d\) is the dimensionality of the data.
        \item \(|\boldsymbol{\Sigma}_j|\) denotes the determinant of the covariance matrix.
        \item \((\cdot)^\top\) represents the transpose of a vector.
    \end{itemize}
    \item \(\mathbf{x}_i\): The \(i\)-th data point in your dataset.
    \item \(\boldsymbol{\mu}_j\): The mean vector of the \(j\)-th Gaussian component.
    \item \(\boldsymbol{\Sigma}_j\): The covariance matrix of the \(j\)-th Gaussian component.
    \item \(k\): The total number of Gaussian components in the mixture model.
    \item \(z_{ij}\): The latent variable indicating the cluster membership of \(\mathbf{x}_i\); \(z_{ij} = 1\) if \(\mathbf{x}_i\) belongs to cluster \(j\), and \(z_{ij} = 0\) otherwise.
    \item \(\sum_{l=1}^{k}\): The summation over all \(k\) Gaussian components.
\end{itemize}

This step assigns soft cluster memberships, meaning that each data point is assigned a probability of belonging to each cluster.

\subsection*{Cluster stability using  Adjusted Mutual Information}

Adjusted Mutual Information (AMI) is a measure used to compare the similarity between two different clusterings of the same dataset. It builds upon the concept of Mutual Information (MI), which quantifies the amount of information shared between two clusterings. MI is able to distinguish between true agreement and agreement that happens purely by chance, by looking at how the knowledge of one set of clusters reduces the surprise at the other set.
However, MI does not account for the similarity that might occur by chance. AMI adjusts the MI score by accounting for the expected similarity between random clusterings, providing a more accurate and reliable metric \cite{JMLR:v11:vinh10a}. 
AMI is symmetric, meaning that the AMI between the two sets is the same regardless of the order of the sets. The AMI used in this paper uses SKlearn's implementation of AMI \cite{Adjusted_mutual_info_score}.

Mutual Information between two clusterings \( U \) and \( V \) is defined as:

\[
\text{MI}(U, V) = \sum_{u \in U} \sum_{v \in V} P(u, v) \log \left( \frac{P(u, v)}{P(u) \, P(v)} \right)
\]

Where:
\begin{itemize}
    \item \( U = \{u_1, u_2, \dots, u_{K_U}\} \) is the set of clusters in clustering \( U \).
    \item \( V = \{v_1, v_2, \dots, v_{K_V}\} \) is the set of clusters in clustering \( V \).
    \item \( P(u) \) is the probability that a randomly selected data point belongs to cluster \( u \) in \( U \).
    \item \( P(v) \) is the probability that a randomly selected data point belongs to cluster \( v \) in \( V \).
    \item \( P(u, v) \) is the joint probability that a data point belongs to cluster \( u \) in \( U \) and cluster \( v \) in \( V \).
\end{itemize}

The AMI adjusts the MI to account for chance, defined as:

\begin{equation}
\label{eq:AMI}
\text{AMI}(U, V) = \frac{\text{MI}(U, V) - \mathbb{E}[\text{MI}(U, V)]}{\text{AVG}[\text{H}(U), \text{H}(V)] - \mathbb{E}[\text{MI}(U, V)]}
\end{equation}

Where:
\begin{itemize}
    \item \( \text{MI}(U, V) \) is the Mutual Information between \( U \) and \( V \).
    \item \( \mathbb{E}[\text{MI}(U, V)] \) is the expected Mutual Information between \( U \) and \( V \) if the cluster assignments were random.
    \item \( \text{H}(U) \) and \( \text{H}(V) \) are the entropies of \( U \) and \( V \), respectively.
    \item \( \text{AVG}[\text{H}(U), \text{H}(V)] \) is the average of the entropies of \( U \) and \( V \).
\end{itemize}

Entropy of a set of clusters measures the uncertainty associated with the cluster assignments:

\[
\text{H}(U) = -\sum_{u \in U} P(u) \log P(u)
\]
\[
\text{H}(V) = -\sum_{v \in V} P(v) \log P(v)
\]

The AMI score ranges from 0 to 1, where close to 1 represents agreement between the two sets and close to 0 represents no relation between the two sets.

\subsection*{Multi-level approaches to clustering}
Previous work has explored multi-level approaches to clustering. One of which uses visualisations called "Clustering Trees" \cite{zappiaClusteringTreesVisualization2018c}. These trees illustrate the proportion of each cluster at a given $K$ that originates from clusters at lower resolutions. By applying this method to synthetic datasets with varying numbers of clusters, the visualisation aids in identifying stable clusters and can help inform the appropriate choice of $K$. Using this method helps determine when additional complexity (in the form of increasing $K$) leads to stable clusters and when further increasing $K$ does not provide meaningful insights.

Another approach to this is the use of MRTree \cite{pengCellTypeHierarchy2021}. Instead of focusing on a visual approach, MRtree uses a more quantifiable method. It determines the optimal number of clusters by analysing stability across different values of $K$ using the Adjusted Rand Index (ARI). Stability is calculated by comparing the clustering results from the reconciled hierarchical tree to the initial non-hierarchical clustering at each resolution. The optimal number of clusters corresponds to the resolution where stability is highest, as indicated by consistent and well-defined partitions.  A change point, where stability drops sharply with further increases in resolution, signifies the threshold for meaningful clustering. In the biological context, this approach generally indicates that the selected number of clusters reflects robust and biologically relevant structures in the data \cite{handlComputationalClusterValidation2005, ronanAvoidingCommonPitfalls2016}. 

\section{Method}

This paper builds on the methodology presented in Miller and Alexander (2025) \cite{Miller2025}, which introduced an approach for human-interpretable clustering of short text using large language models. Specifically, this work adopts the framework found in the earlier work for text embedding and clustering, applying it to analyze the movement of Twitter user biographies (bios) between clusters as the number of clusters increased from 1 to 20.

Our analysis involves clustering bios using Gaussian Mixture Models (GMM), generating cluster names with the Google Gemini language model, extracting significant keywords for each cluster, and visualizing the transitions using a Sankey diagram. By leveraging the foundation established in \cite{Miller2025}, we extend the methodology to explore how clusters evolve as the number of components increases, providing new insights into the interpretability of clustering results.

\subsection*{Clustering}

Twitter bios were collected and preprocessed by converting emojis into text descriptions using the \texttt{emoji} library \cite{2022emoji}.  Embeddings for each bio were created using MiniLM \cite{wang2020}, representing the bios in a high-dimensional vector space suitable for clustering.

Clustering was performed using Gaussian Mixture Models (GMM) from the \texttt{scikit-learn} library. The number of clusters $k$ was varied from 1 to 20 to observe how cluster assignments changed with increasing cluster numbers. For each value of $k$, a GMM was fitted with diagonal covariance, a maximum of 2000 iterations, and a random state of 0 
to ensure reproducibility. The GMM assigned each bio to one of the $k$ clusters based on the embeddings, resulting in a mapping of bios to clusters across different values of $k$.

\subsection*{Cluster Naming}

To generate descriptive names for each cluster, the Google Gemini language model was utilized. For each cluster at each value of $k$, a prompt was created that included the top 10 most frequent words in the cluster and a random sample of 20 bios from that cluster. The prompt was structured as follows:

\begin{quote}
``Create a name for the following cluster of Twitter bios. It has the following top 10 most frequent words: \newline
\textit{word1, word2, word3, \dots, word10} \newline
And this is a random sample of Twitter bios from the cluster:
\newline
1. \textit{Bio1} \newline
2. \textit{Bio2} \newline
3. \textit{Bio3} \newline
\dots \newline
20. \textit{Bio20}''
\end{quote}

The prompt was sent to the Google Gemini model via its API, requesting a concise and descriptive name for the cluster. To handle potential non-unique names (as the model might generate the same name for different clusters), uniqueness was ensured by appending numbers to duplicate names. This method allowed differentiation of clusters with identical names without altering their appearance in the visualisation.

\subsection*{Cluster Visualisation}

A Sankey diagram was constructed using the Plotly Python library to visualize the movement of bios between clusters as the number of clusters increased from 1 to 20. For each bio, cluster assignments across values of $k$ from 1 to 20 were tracked. Edges between clusters at consecutive values of $k$ ($k$ and $k+1$) were created based on how many bios transitioned between a cluster at $k$ and $k + 1$. Edges representing fewer than 150 bios were filtered out to focus on significant transitions. Each node in the diagram represented a cluster at a specific value of $k$, labeled with the unique cluster names generated by the Google Gemini model. Nodes were colored based on cluster stability (see Eq.~(\ref{eq:Stability}) for details as to how stability is calculated).

\subsection*{Stability Assessment}

The first stability assessment aimed to determine the stability of the clustering algorithm to the selection of dimensions within the embedding space.  The goal of this assessment was to identify if any dimensions in MiniLM are essential, or if the loss of dimensions impacted the clustering. The embeddings were based on the MiniLM model, which provides 384-dimensional representations of the bios. For this analysis, an 80\% random subsample of the dimensions (approximately 307 dimensions) was selected without replacement from the full set of 384 dimensions.

Clustering was performed on these reduced-dimensionality embeddings using the same Gaussian Mixture Model (GMM) configuration as described previously, varying the number of clusters $k$ from 1 to 20. The Adjusted Mutual Information (AMI) score was calculated between the clusters obtained from the subsampled dimensions and the original clusters derived from the full set of dimensions. 
The AMI is a measure of the agreement between two clusterings, adjusted for chance, with a value of 1 indicating perfect agreement and a value close to 0 indicating agreement no better than random.

This resampling and clustering process was repeated 100 times, each time selecting a different random 80\% subset of dimensions. The distribution of AMI scores across the iterations provided an indication of the stability of the clustering results with respect to variations in the feature space.

The second assessment focused on the robustness of the clustering results to variations in the data sample. An 80\% random subsample of the bios was selected without replacement from the full dataset. Clustering was performed on this subset using the full 384-dimensional embeddings and the same GMM configuration.

The AMI score was calculated between the clusters obtained from the 80\% bios sample and the clusters obtained from the full dataset. This process was repeated 100 times with different random samples of bios. The resulting distribution of AMI scores indicated the extent to which the clustering results were consistent across different samples of the data.

The third assessment investigated the effect of random initialisation on the clustering results. The GMM clustering algorithm involves stochastic elements that can lead to different results depending on the random seed used for initialization. Clustering was performed using different random seeds, specifically seeds ranging from 1 to 100, while keeping all other parameters constant.

For each seed, the AMI score was calculated between the clusters obtained with that seed and the clusters obtained with the initial seed (seed = 0). Analyzing the AMI scores across different seeds provided insight into the sensitivity of the clustering outcomes to random initialization and whether certain cluster structures were consistently identified regardless of the seed.

To measure the stability accross clusters we introduce the measure called "Proportional Stability" which is given by:
\begin{equation}
\text{Proportional Stability} = \frac{1}{K} \sum_{k=1}^{K} \frac{\displaystyle \max_{k'} \left| C_k \cap C'_{k'} \right|}{\left| C_k \right|}
\label{eq:Stability}
\end{equation}

\noindent
\textbf{Where:}

\begin{itemize}
    \item $K$ is the number of clusters in the current model.
    \item $C_k$ is the set of data points in cluster $k$ of the current model.
    \item $C'_{k'}$ is the set of data points in cluster $k'$ of the previous model.
    \item $\left| C_k \right|$ denotes the number of data points in cluster $C_k$.
    \item $\left| C_k \cap C'_{k'} \right|$ is the number of data points common to both $C_k$ and $C'_{k'}$.
    \item $\displaystyle \max_{k'} \left| C_k \cap C'_{k'} \right|$ finds the previous cluster $C'_{k'}$ that contributes the maximum number of data points to $C_k$.
\end{itemize}

\section{{Results}}
In the following results, we first examine the single-level stability of the Gaussian Mixture Model (GMM)-based clustering of Twitter user bios across varying numbers of clusters ($K$). Single-level stability refers to the robustness of the clustering solution at a fixed number of clusters. We then consider the multi-level stability, which focuses on how cluster structures evolve as $K$ increases. To investigate this, we visualise how user bios transition between clusters at successive resolutions, and characterise these clusters using descriptive names generated by the Google Gemini language model.
\subsection*{Single-Level Stability}

\begin{figure}[htbp]
 \begin{flushleft}
  \includegraphics[width=0.9\linewidth]{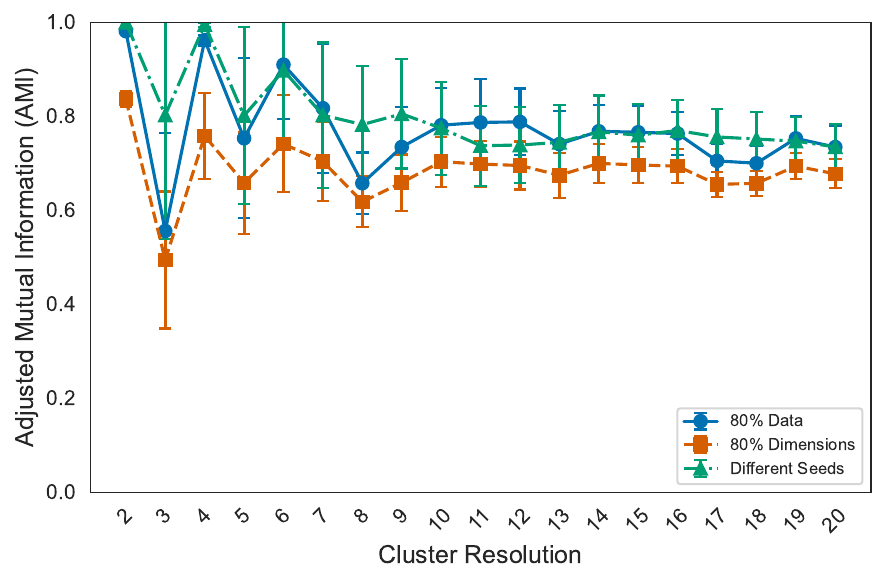}
  \caption{The average AMI (\ref{eq:AMI}) between the original clustering (seed = 0), and 100 iterations of clusters created using only 80\% of dimensions of the embedding space created by the MiniLM Language Model, 80\% of the Data, and different seeds. The error bars represent the standard deviation of the AMI.}
  \label{fig:Stability_Plot_Error}
  \end{flushleft}
\end{figure}

Figure \ref{fig:Stability_Plot_Error} presents results on clustering stability, through the average Adjusted Mutual Information (AMI) score Eq.~(\ref{eq:AMI}), under three conditions: subsetting dimensions, subsetting data, and varying random seeds. The AMI remains relatively stable across all three measures. Although there is a minor fluctuation at three clusters, where stability dips slightly for column and data resampling, this difference is within one standard deviation of other $K$ values and is therefore not statistically notable. Beyond 10 clusters, the AMI shows minimal fluctuation across all three metrics, indicating that the clustering configuration is relatively robust at higher $K$ values.
From a practical standpoint, these results suggest no strongly unsuitable values of $K$ based solely on stability. High stability means that, under the tested conditions, the clusters remain structurally consistent even when the feature space or initialisation changes. The stability analysis thus provides a method to exclude certain $K$ values, identifying clusters that are less sensitive to perturbations and therefore more likely to represent genuine, consistent patterns in the data, rather than clusters arising from random noise.

\subsection*{Multi-Level Stability}
Although Figure \ref{fig:Stability_Plot_Error} offers insight into the stability of each individual $K$, it does not reveal how clusters evolve as $K$ changes. Multi-level stability examines whether increasing $K$ produces new clusters that are entirely distinct, or whether they emerge from the subdivision of existing clusters. Understanding this evolution is essential, as it allows analysts to discern whether cluster structures remain coherent as granularity increases, or whether cluster identities shift substantially.

\begin{figure}[htbp]
 \begin{flushleft}
  \includegraphics[width=0.9\linewidth]{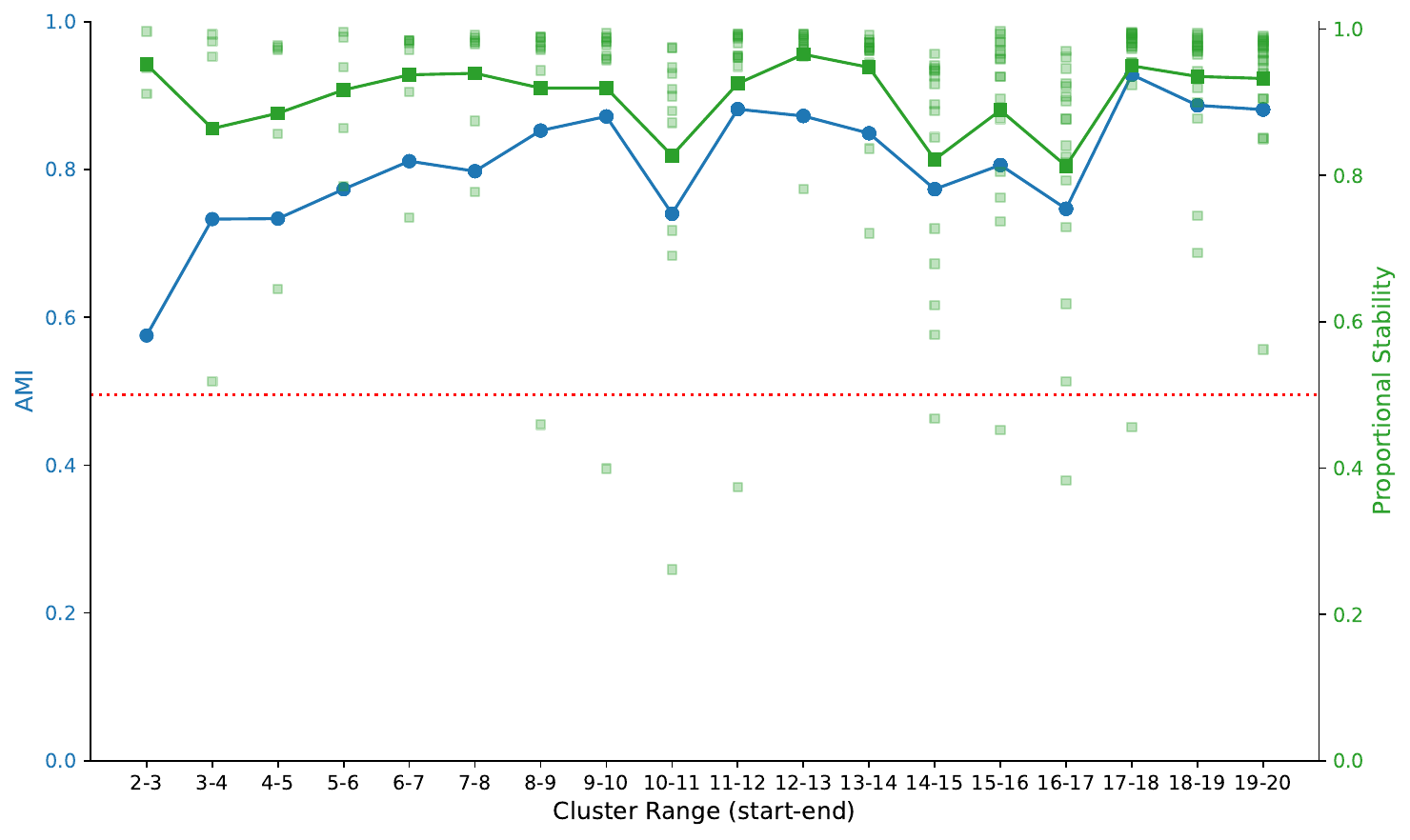}
  \caption[Proportional Stability]{
   The AMI (\ref{eq:AMI}) (blue line, circles) and proportional stability (\ref{eq:Stability}) (green line, squares) are shown between successive clustering levels, as $K$ increases.  The scattered green circles show the individual proportional stability results for each cluster.  A proportional stability close to 1 indicates that a cluster has largely a single parent.  The red dashed line corresponds to a proportional stability of 0.5.  Clusters sitting below this line are `new' in that they are combinations of clusters at the lower resolution.}
  \label{fig:Between Clusters}
  \end{flushleft}
\end{figure}

Figure \ref{fig:Between Clusters} shows the AMI between consecutive models as $K$ increases, along with the average Proportional Stability—a measure of how closely each cluster at the new level corresponds to one or more clusters at the previous level. The AMI generally hovers around 0.7–0.8, indicating that new clusters typically form by splitting from existing ones, rather than representing entirely novel structures. Similarly, Proportional Stability remains above 0.8 for most $K$ values, suggesting that clusters retain a substantial portion of their membership as the clustering resolution becomes finer. Across all cluster transitions from $K= 2$ to $K= 20$
K=20, only 8 clusters exhibit a proportional stability below 0.5. This indicates that the vast majority of clusters are primarily derived from a single dominant cluster at the previous level.
It should be emphasised that when changing $K$ the clustering is carried out without any information about the clustering results at the different $K$ value.  This implies that the overarching cluster solutions appear to remain similar across consecutive values of $K$. However, it is important to examine individual clusters to fully understand their behaviour, as some clusters may remain stable throughout, reflecting cohesive user communities, while others may fragment or recombine as $K$ grows.

\begin{figure}[ht]
\centering
\includegraphics[width=\linewidth]{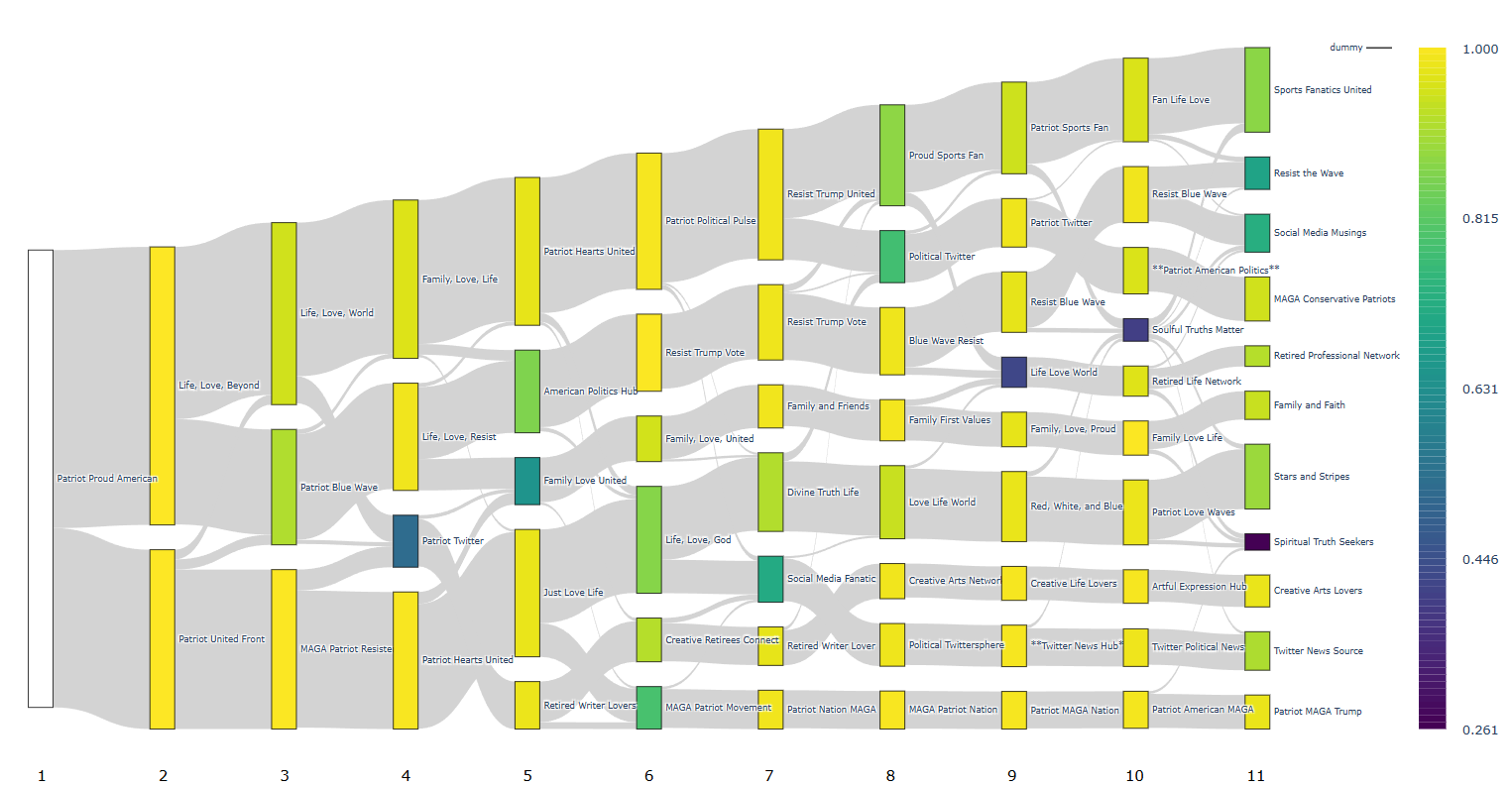}
\caption{Shows the proportion of bios that can be found in a single cluster at the previous hierarchical clustering level 
as the number of clusters increases from 1 - 11.  Names are created using Google Gemini and a sample of bios from the cluster and the top words. The colour of each cluster is given by the Proportional Stability of the cluster. Where the more yellow (lighter) a cluster is the greater proportion of it came from a cluster at a previous resolution. Whereas the more blue (darker) a cluster is, the more the cluster is made up of a mix of different clusters at a lower resolution.}
\label{fig:Sankey10}
\end{figure}

Figure \ref{fig:Sankey10} provides a more granular view of multi-level stability. Initially, the clustering separates bios into broad categories, such as political versus non-political. As $K$ increases, new clusters generally emerge as subdivisions of existing clusters rather than aggregations of disparate segments. This pattern suggests that some thematic groupings remain coherent even as the clustering resolution increases.
These stable memberships are often reflected in the cluster names generated by Gemini. For example, from $K=6$ to $K=11$, a stable set of clusters consistently include terms such as "MAGA" and "Patriot." This consistency in both membership and linguistic characterisation underscores the presence of cohesive user communities, possibly reflecting strongly aligned political identities. In contrast, certain clusters such as "Soulful Truths Matter" show lower Proportional Stability (as low as 0.261) and fragment into multiple smaller clusters at higher resolutions, showing that not all clusters are equally persistent as $K$ changes.
Nonetheless, such visualisations, when combined with expert or domain-specific insights, can guide the choice of appropriate $K$ values for more detailed analysis. Increasing $K$ inevitably raises cognitive load, as more clusters require greater interpretative effort. Thus, selecting a lower $K$, when clusters of interest first appear, may be preferable when the goal is to derive meaningful insights without overcomplicating the analysis.

\section{Discussion}
The stability metrics, including Adjusted Mutual Information (AMI) and Proportional Stability, consistently showed high values (generally above 0.7 and 0.8, respectively), reinforcing the observation that the clusters are robust to changes in $K$. Notably, the AMI between successive clusterings remained relatively constant, suggesting that the fundamental structure of the data is being preserved as the number of clusters increases. There are, of course, a few exceptions within each resolution, where new clusters form from many other clusters, and likewise some clusters will splinter into multiple other clusters with no clear successor at a higher resolution. However, on the whole, most clusters are able to trace their lineage through many levels of clustering.


This stability implies that the clustering algorithm is capturing intrinsic clusters in the data that are resilient to the choice of \(K\). This is somewhat surprising because, as shown in equation~\eqref{eq:e-step}, the responsibilities \(\gamma(z_{ij})\) that assign data points to clusters depend on \(K\) through the summation over all \(k\) components in the denominator. Changing \(K\) alters the number of Gaussian components and their parameters \(\pi_j\), \(\boldsymbol{\mu}_j\), and \(\boldsymbol{\Sigma}_j\), which are re-estimated during the EM algorithm. Since these parameters can vary significantly with different values of \(K\), there is no guarantee that similar clusters will appear when \(K\) changes.

An interesting observation was that the AMI for all three stability metrics remained fairly constant at higher values of \(K\) (Figure~\ref{fig:Between Clusters}). This may be a result of how, as \(K\) increases, the relative percentage increase in the number of clusters becomes smaller, leading to less drastic changes between clusterings.

The stability assessments with respect to random initialisations, resampling data, and resampling dimensions demonstrated that the clustering results are robust to these perturbations (Figure~\ref{fig:Stability_Plot_Error}). The AMI scores remained high across different seeds, subsets of data, and subsets of dimensions, indicating that the clustering algorithm consistently identifies similar structures in the data. This robustness is essential for practical applications, as it suggests that the clustering results are reliable and not unduly influenced by random factors.

Nevertheless, several limitations should be acknowledged. First, the study relied on a single clustering algorithm (GMM), and while it is well-suited for data that may be modelled as a mixture of Gaussians, other algorithms may capture different aspects of the data structure. Second, the embeddings were generated using a specific model (MiniLM), and different embedding techniques may yield different results. Third, the use of the Google Gemini language model for cluster naming introduces potential biases and may not always produce accurate or unique names, as the model's outputs are influenced by its training data and prompt design. Additionally, this method was only tested on one dataset. The stability observed may be a feature of this particular dataset rather than the method itself. Future testing on other datasets is necessary to ensure reliability.


\section{Conclusion}

This study assessed the stability of clustering short-text data using a combination of Gaussian Mixture Models (GMM) and embeddings generated by a large language model (LLM). Specifically, we examined how clustering results evolve internally and across different values of $K$. Our findings demonstrate that clustering is stable across resolutions, with clusters primarily subdividing rather than reorganizing as $K$ increases. This pattern was visualized in the Sankey diagram (Figure~\ref{fig:Sankey10}), where most data points remained within the same cluster lineage, underscoring that the algorithm captures intrinsic structures that are robust to changes in $K$.

These results challenge the traditional focus on identifying a single ``optimal” number of clusters. Instead, they advocate for examining the interpretability and utility of clusters across multiple resolutions. Visualisations such as Sankey diagrams allow users to explore the relationships and transitions between clusters, enabling decisions informed by practical utility rather than abstract metrics.

Future work should expand on this approach by applying it to different datasets and clustering methods to assess generalisability. Incorporating human evaluations to assess the semantic coherence and usability of clusters would also provide valuable insights. Ultimately, this study underscores the importance of prioritizing user-centered clustering solutions that balance stability, interpretability, and visual accessibility, moving the focus from what is ``optimal” to what is most useful.


\bibliography{Main_File}

\begin{thebibliography}{10}
\providecommand{\url}[1]{#1}
\csname url@samestyle\endcsname
\providecommand{\newblock}{\relax}
\providecommand{\bibinfo}[2]{#2}
\providecommand{\BIBentrySTDinterwordspacing}{\spaceskip=0pt\relax}
\providecommand{\BIBentryALTinterwordstretchfactor}{4}
\providecommand{\BIBentryALTinterwordspacing}{\spaceskip=\fontdimen2\font plus
\BIBentryALTinterwordstretchfactor\fontdimen3\font minus \fontdimen4\font\relax}
\providecommand{\BIBforeignlanguage}[2]{{%
\expandafter\ifx\csname l@#1\endcsname\relax
\typeout{** WARNING: IEEEtran.bst: No hyphenation pattern has been}%
\typeout{** loaded for the language `#1'. Using the pattern for}%
\typeout{** the default language instead.}%
\else
\language=\csname l@#1\endcsname
\fi
#2}}
\providecommand{\BIBdecl}{\relax}
\BIBdecl

\bibitem{van2023white}
\BIBentryALTinterwordspacing
I.~V. Mechelen, A.-L. Boulesteix, R.~Dangl, N.~Dean, C.~Hennig, F.~Leisch, D.~Steinley, and M.~J. Warrens, ``A white paper on good research practices in benchmarking: The case of cluster analysis,'' \emph{Wiley Interdisciplinary Reviews: Data Mining and Knowledge Discovery}, vol.~13, no.~6, p. e1511, 2023. [Online]. Available: \url{https://doi.org/10.1002/widm.1511}
\BIBentrySTDinterwordspacing

\bibitem{roschPrinciplesCategorization1978}
\BIBentryALTinterwordspacing
E.~Rosch, ``Principles of categorization,'' in \emph{Cognition and Categorization}, E.~Rosch and B.~B. Lloyd, Eds.\hskip 1em plus 0.5em minus 0.4em\relax Hillsdale, NJ: Lawrence Erlbaum Associates, 1978, pp. 27--48. [Online]. Available: \url{https://escholarship.org/uc/item/0sz9c8qh}
\BIBentrySTDinterwordspacing

\bibitem{hennig2019cluster}
C.~Hennig, ``Cluster validation by measurement of clustering characteristics relevant to the user,'' in \emph{Data Analysis and Applications 1: Clustering and Regression, Modeling--Estimating, Forecasting and Data Mining}, C.~Gatu, E.~Diday, and G.~Saporta, Eds.\hskip 1em plus 0.5em minus 0.4em\relax Hoboken, NJ: John Wiley \& Sons, 2019, vol.~2, pp. 1--24.

\bibitem{akhanliComparingClusteringsNumbers2020}
\BIBentryALTinterwordspacing
S.~E. Akhanli and C.~Hennig, ``Comparing clusterings and numbers of clusters by aggregation of calibrated clustering validity indexes,'' \emph{Statistics and Computing}, vol.~30, no.~5, pp. 1523--1544, 2020. [Online]. Available: \url{https://doi.org/10.1007/s11222-020-09958-2}
\BIBentrySTDinterwordspacing

\bibitem{grimmer2021machine}
J.~Grimmer, M.~E. Roberts, and B.~M. Stewart, ``Machine learning for social science: An agnostic approach,'' \emph{Annual Review of Political Science}, vol.~24, pp. 395--419, 2021.

\bibitem{pedryczFuzzyClusteringKnowledgebased2004}
\BIBentryALTinterwordspacing
W.~Pedrycz, ``Fuzzy clustering with a knowledge-based guidance,'' \emph{Pattern Recognition Letters}, vol.~25, no.~4, pp. 469--480, 2004. [Online]. Available: \url{https://www.sciencedirect.com/science/article/pii/S0167865503002770}
\BIBentrySTDinterwordspacing

\bibitem{lakhawatNovelClusteringAlgorithm2016}
\BIBentryALTinterwordspacing
P.~Lakhawat, M.~Mishra, and A.~Somani, ``A novel clustering algorithm to capture utility information in transactional data,'' in \emph{Proceedings of the 8th International Joint Conference on Knowledge Discovery, Knowledge Engineering and Knowledge Management}.\hskip 1em plus 0.5em minus 0.4em\relax SCITEPRESS - Science and Technology Publications, 2016, pp. 456--462. [Online]. Available: \url{http://www.scitepress.org/DigitalLibrary/Link.aspx?doi=10.5220/0006092104560462}
\BIBentrySTDinterwordspacing

\bibitem{krausMultiobjectiveSelectionCollecting2011}
\BIBentryALTinterwordspacing
J.~M. Kraus, C.~M{\"u}ssel, G.~Palm, and H.~A. Kestler, ``Multi-objective selection for collecting cluster alternatives,'' \emph{Computational Statistics}, vol.~26, no.~2, pp. 341--353, 2011. [Online]. Available: \url{https://doi.org/10.1007/s00180-011-0244-6}
\BIBentrySTDinterwordspacing

\bibitem{krishnapuramPossibilisticApproachClustering1993}
\BIBentryALTinterwordspacing
R.~Krishnapuram and J.~Keller, ``A possibilistic approach to clustering,'' \emph{IEEE Transactions on Fuzzy Systems}, vol.~1, no.~2, pp. 98--110, 1993. [Online]. Available: \url{https://ieeexplore.ieee.org/document/227387}
\BIBentrySTDinterwordspacing

\bibitem{stevensExploringTopicCoherence2012}
\BIBentryALTinterwordspacing
K.~Stevens, P.~Kegelmeyer, D.~Andrzejewski, and D.~Buttler, ``Exploring topic coherence over many models and many topics,'' in \emph{Proceedings of the 2012 Joint Conference on Empirical Methods in Natural Language Processing and Computational Natural Language Learning}.\hskip 1em plus 0.5em minus 0.4em\relax Jeju Island, Korea: Association for Computational Linguistics, 2012, pp. 952--961. [Online]. Available: \url{https://aclanthology.org/D12-1087}
\BIBentrySTDinterwordspacing

\bibitem{rousseeuwSilhouettesGraphicalAid1987}
\BIBentryALTinterwordspacing
P.~J. Rousseeuw, ``Silhouettes: A graphical aid to the interpretation and validation of cluster analysis,'' \emph{Journal of Computational and Applied Mathematics}, vol.~20, pp. 53--65, 1987. [Online]. Available: \url{https://www.sciencedirect.com/science/article/pii/0377042787901257}
\BIBentrySTDinterwordspacing

\bibitem{ahmedKmeansAlgorithmComprehensive2020}
\BIBentryALTinterwordspacing
M.~Ahmed, R.~Seraj, and S.~M.~S. Islam, ``The k-means algorithm: A comprehensive survey and performance evaluation,'' \emph{Electronics}, vol.~9, no.~8, p. 1295, 2020. [Online]. Available: \url{https://www.mdpi.com/2079-9292/9/8/1295}
\BIBentrySTDinterwordspacing

\bibitem{hosseiniAlternativeEMGaussian2020}
\BIBentryALTinterwordspacing
R.~Hosseini and S.~Sra, ``An alternative to em for gaussian mixture models: Batch and stochastic riemannian optimization,'' \emph{Mathematical Programming}, vol. 181, no.~1, pp. 187--223, 2020. [Online]. Available: \url{https://doi.org/10.1007/s10107-019-01381-4}
\BIBentrySTDinterwordspacing

\bibitem{kiselevChallengesUnsupervisedClustering2019}
\BIBentryALTinterwordspacing
V.~Y. Kiselev, T.~S. Andrews, and M.~Hemberg, ``Challenges in unsupervised clustering of single-cell rna-seq data,'' \emph{Nature Reviews Genetics}, vol.~20, no.~5, pp. 273--282, 2019. [Online]. Available: \url{https://www.nature.com/articles/s41576-018-0088-9}
\BIBentrySTDinterwordspacing

\bibitem{handlComputationalClusterValidation2005}
\BIBentryALTinterwordspacing
J.~Handl, J.~Knowles, and D.~B. Kell, ``Computational cluster validation in post-genomic data analysis,'' \emph{Bioinformatics}, vol.~21, no.~15, pp. 3201--3212, 2005. [Online]. Available: \url{https://doi.org/10.1093/bioinformatics/bti517}
\BIBentrySTDinterwordspacing

\bibitem{ronanAvoidingCommonPitfalls2016}
\BIBentryALTinterwordspacing
T.~Ronan, Z.~Qi, and K.~M. Naegle, ``Avoiding common pitfalls when clustering biological data,'' \emph{Science Signaling}, vol.~9, no. 432, pp. re6--re6, 2016. [Online]. Available: \url{https://www.science.org/doi/10.1126/scisignal.aad1932}
\BIBentrySTDinterwordspacing

\bibitem{yu2022benchmarking}
L.~Yu, Y.~Cao, J.~Y.~H. Yang, and P.~Yang, ``Benchmarking clustering algorithms on estimating the number of cell types from single-cell rna-sequencing data,'' \emph{Genome Biology}, vol.~23, no.~1, p.~49, 2022.

\bibitem{qiClusteringClassificationMethods2020}
\BIBentryALTinterwordspacing
R.~Qi, A.~Ma, Q.~Ma, and Q.~Zou, ``Clustering and classification methods for single-cell rna-sequencing data,'' \emph{Briefings in Bioinformatics}, vol.~21, no.~4, pp. 1196--1208, 2020. [Online]. Available: \url{https://doi.org/10.1093/bib/bbz062}
\BIBentrySTDinterwordspacing

\bibitem{JMLR:v18:17-069}
\BIBentryALTinterwordspacing
F.~Morstatter and H.~Liu, ``In search of coherence and consensus: Measuring the interpretability of statistical topics,'' \emph{Journal of Machine Learning Research}, vol.~18, no. 169, pp. 1--32, 2018. [Online]. Available: \url{http://jmlr.org/papers/v18/17-069.html}
\BIBentrySTDinterwordspacing

\bibitem{kuncheva2006evaluation}
L.~I. Kuncheva and D.~P. Vetrov, ``Evaluation of stability of k-means cluster ensembles with respect to random initialization,'' \emph{IEEE Transactions on Pattern Analysis and Machine Intelligence}, vol.~28, no.~11, pp. 1798--1808, 2006.

\bibitem{lord2017using}
E.~Lord, M.~Willems, F.-J. Lapointe, and V.~Makarenkov, ``Using the stability of objects to determine the number of clusters in datasets,'' \emph{Information Sciences}, vol. 393, pp. 29--46, 2017.

\bibitem{21GaussianMixture}
\BIBentryALTinterwordspacing
(2024) 2.1. gaussian mixture models. scikit-learn. [Online]. Available: \url{https://scikit-learn/stable/modules/mixture.html}
\BIBentrySTDinterwordspacing

\bibitem{mit_ml_notes_2015}
\BIBentryALTinterwordspacing
{Massachusetts Institute of Technology}, ``Algorithmic aspects of machine learning: Chapter 6 notes,'' 2015. [Online]. Available: \url{https://ocw.mit.edu/courses/18-409-algorithmic-aspects-of-machine-learning-spring-2015/e339520c4069ca5e785b29a3c604470e\_MIT18\_409S15\_chapp6.pdf}
\BIBentrySTDinterwordspacing

\bibitem{JMLR:v25:23-1245}
\BIBentryALTinterwordspacing
X.~Li, J.~Zhou, and H.~Wang, ``Gaussian mixture models with rare events,'' \emph{Journal of Machine Learning Research}, vol.~25, no. 252, pp. 1--40, 2024. [Online]. Available: \url{http://jmlr.org/papers/v25/23-1245.html}
\BIBentrySTDinterwordspacing

\bibitem{ghosh2018emnotes}
\BIBentryALTinterwordspacing
S.~Mukherjee, ``Lecture notes on expectation-maximization algorithm,'' 2018. [Online]. Available: \url{http://www2.stat.duke.edu/~sayan/Sta613/2018/lec/emnotes.pdf}
\BIBentrySTDinterwordspacing

\bibitem{JMLR:v11:vinh10a}
\BIBentryALTinterwordspacing
N.~X. Vinh, J.~Epps, and J.~Bailey, ``Information theoretic measures for clusterings comparison: Variants, properties, normalization and correction for chance,'' \emph{Journal of Machine Learning Research}, vol.~11, no.~95, pp. 2837--2854, 2010. [Online]. Available: \url{http://jmlr.org/papers/v11/vinh10a.html}
\BIBentrySTDinterwordspacing

\bibitem{Adjusted_mutual_info_score}
\BIBentryALTinterwordspacing
(2024) Adjusted mutual info score. scikit-learn. [Online]. Available: \url{https://scikit-learn/stable/modules/generated/sklearn.metrics.adjusted\_mutual\_info\_score.html}
\BIBentrySTDinterwordspacing

\bibitem{zappiaClusteringTreesVisualization2018c}
L.~Zappia and A.~Oshlack, ``Clustering trees: A visualization for evaluating clusterings at multiple resolutions,'' \emph{GigaScience}, vol.~7, no.~7, p. giy083, 2018.

\bibitem{pengCellTypeHierarchy2021}
\BIBentryALTinterwordspacing
M.~Peng, B.~Wamsley, A.~G. Elkins, D.~H. Geschwind, Y.~Wei, and K.~Roeder, ``Cell type hierarchy reconstruction via reconciliation of multi-resolution cluster tree,'' \emph{Nucleic Acids Research}, vol.~49, no.~16, p. e91, 2021. [Online]. Available: \url{https://doi.org/10.1093/nar/gkab481}
\BIBentrySTDinterwordspacing

\bibitem{Miller2025}
J.~K. Miller and T.~J. Alexander, ``Human-interpretable clustering of short text using large language models,'' \emph{Royal Society Open Science}, vol.~12, p. 241692, 2025.

\bibitem{2022emoji}
\BIBentryALTinterwordspacing
T.~Kim, K.~Wurster, and T.~Jalilov, ``emoji: Emoji for python,'' 2022, version 2.0.0. [Online]. Available: \url{https://pypi.org/project/emoji/}
\BIBentrySTDinterwordspacing

\bibitem{wang2020}
\BIBentryALTinterwordspacing
W.~Wang, F.~Wei, L.~Dong, H.~Bao, N.~Yang, and M.~Zhou, ``Minilm: Deep self-attention distillation for task-agnostic compression of pre-trained transformers,'' in \emph{Proceedings of the 34th International Conference on Neural Information Processing Systems (NeurIPS)}.\hskip 1em plus 0.5em minus 0.4em\relax Curran Associates Inc., 2020, pp. 5776--5788. [Online]. Available: \url{https://proceedings.neurips.cc/paper\_files/paper/2020/file/3f5ee243547dee91fbd053c1c4a845aa-Paper.pdf}
\BIBentrySTDinterwordspacing

\end{thebibliography}
\end{document}